\documentclass[conference]{IEEEtran}
\IEEEoverridecommandlockouts

\usepackage{cite}
\usepackage{amsmath,amssymb,amsfonts}
\usepackage{graphicx}
\usepackage{algpseudocode}
\usepackage{textcomp}
\usepackage{xcolor}
\usepackage{url}
\usepackage{multirow}
\usepackage{array}
\usepackage[protrusion=false]{microtype}

\usepackage[letterpaper,hmargin=1in,vmargin=1.04in,includefoot,footskip=24pt]{geometry}
\makeatletter
\renewcommand{\small}{\@setfontsize\small{10}{12}}
\renewcommand{\footnotesize}{\@setfontsize\footnotesize{10}{12}}
\makeatother
\def\BibTeX{{\rm B\kern-.05em{\sc i\kern-.025em b}\kern-.08em
    T\kern-.1667em\lower.7ex\hbox{E}\kern-.125emX}}

\usepackage[hidelinks]{hyperref}
\usepackage{cleveref}
\newcounter{PMRalgorithm}

\crefname{figure}{Fig.}{Figs.}
\crefname{table}{Table}{Tables}
\crefname{equation}{Eq.}{Eqs.}
\crefname{section}{Sec.}{Secs.}
\crefname{algorithm}{Algorithm}{Algorithms}

\begin{document}

\title{\fontsize{17}{22}\selectfont PMR: MoE Routing with Shared Prototypes and Congestion Pricing
\thanks{\textsuperscript{*}Corresponding authors: Lei Duan, Yuqi Ouyang, and Dengchun Li.}
}

\author{%
\IEEEauthorblockN{\normalsize \makebox[\textwidth][l]{Wending Wang\textsuperscript{1},\enspace Binqi Huang\textsuperscript{2},\enspace Ju'an Li\textsuperscript{1},\enspace Yu Guo\textsuperscript{1},\enspace Lei Duan\textsuperscript{1,*},\enspace Yuqi Ouyang\textsuperscript{1,*},\enspace Dengchun Li\textsuperscript{1,*}}}
\IEEEauthorblockA{\normalfont
\makebox[\textwidth][l]{\textsuperscript{1}Sichuan University, Chengdu, China}\\
\makebox[\textwidth][l]{\textsuperscript{2}University of Science and Technology of China, Hefei, China}\\
\parbox[t]{\textwidth}{\raggedright \mbox{\nolinkurl{wangwending@stu.scu.edu.cn}}, \mbox{\nolinkurl{huangbinqi0038@mail.ustc.edu.cn}}, \mbox{\nolinkurl{lijuan337@outlook.com}}, \mbox{\nolinkurl{guoyugy@stu.scu.edu.cn}}, \mbox{\nolinkurl{leiduan@scu.edu.cn}}, \mbox{\nolinkurl{yuqi.ouyang@scu.edu.cn}}, \mbox{\nolinkurl{mikecovlee@163.com}}}}
}

\maketitle
\thispagestyle{plain}

\begin{abstract}
Conventional linear routers in sparse mixture-of-experts models select experts directly from token–expert matching scores, coupling input matching with expert allocation. We propose Prototype-Mediated Routing (PMR), a two-stage router with shared prototypes. PMR first computes each token’s distribution over prototypes, then uses it to mix prototype-conditioned expert distributions according to the law of total probability. This formulation models input matching and expert allocation separately, explicitly represents many-to-many associations between prototypes and experts, and provides separate interfaces for context adaptation and load regulation. Within this framework, short-term states summarize preceding tokens to adapt the prototypes, while congestion prices updated from dispatch feedback achieve loss-free load balancing by adjusting the conditional expert distributions. We use a two-pass blockwise approximation to accelerate short-term state processing, achieving over 20 times the training throughput of the eager tokenwise PMR implementation while maintaining high agreement in routing decisions. Across three pretraining configurations with 0.43B, 1.18B, and 0.48B total parameters, PMR improves mean zero-shot accuracy on seven benchmarks over linear routing baselines with matched expert architectures by 1.54, 2.65, and 1.45 percentage points (pp), respectively. The corresponding gains on ARC-Easy are 6.71, 5.80, and 3.92~pp. In parameter-efficient fine-tuning of 7B--8B backbones, PMR's improvements over MixLoRA vary across backbones, with gains of up to 1.76~pp in mean accuracy across four tasks.
\end{abstract}

\section{Introduction}
Mixture-of-Experts (MoE) models expand language model capacity without a proportional increase in per-token computation by activating only a small subset of feed-forward network (FFN) experts for each token~\cite{shazeer2017outrageously,lepikhin2021gshard,fedus2022switch,jiang2024mixtral}. For instance, Qwen3-30B-A3B contains 30.5 billion total parameters but activates only 3.3 billion parameters per token during inference~\cite{yang2025qwen3}. This efficiency, however, depends critically on the router, which determines how input tokens are distributed among experts. Beyond controlling computational allocation, routing decisions also shape the training data received by individual experts and, consequently, their specialization. Accordingly, considerable research has focused on improving expert utilization and preventing imbalanced token distribution through auxiliary load-balancing objectives and related mechanisms~\cite{shazeer2017outrageously,lepikhin2021gshard,fedus2022switch}. These approaches regulate aggregate expert utilization ~\cite{qiu2025,wang2024auxiliary}. At the level of individual assignments, however, conventional linear routing leaves the mapping from input patterns to experts implicit in its scores.

\begin{figure}[t]
    \centering
    \includegraphics[width=\columnwidth]{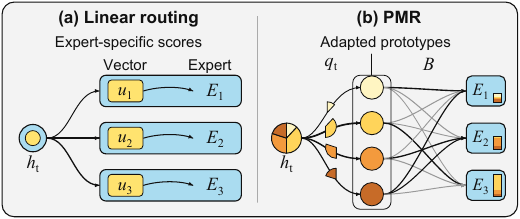}
    \caption{(a) Linear routing computes expert-indexed scores from the token representation. (b) PMR separately represents the prototype response distribution ($q_t$) and the prototype--expert association table ($B$). Token fragments illustrate matching responses. The selected experts process the full token representation.}
    \label{fig:routing-motivation}
\end{figure}

To examine how input features relate to expert preferences, we first analyze the scoring process of conventional linear routing. As illustrated in \cref{fig:routing-motivation}(a), a linear router maintains one learnable routing vector per expert and computes expert scores as inner products between the input token representation and these vectors~\cite{shazeer2017outrageously}. It then selects experts directly from these scores, without separately representing which input patterns the token matches and how those patterns relate to experts. Findings from RouterInterp show that a single expert can be associated with multiple semantically distinct input features, while the same feature may be associated with multiple experts~\cite{lasy2026routerinterp}. This observation motivates using shared matching references without preassigning them to individual experts, and modeling their expert associations separately, as illustrated in \cref{fig:routing-motivation}(b). This separation allows sequence context to adapt the matching references while dispatch feedback adjusts expert allocation. In this work, we investigate whether this design improves downstream performance at a fixed number of active experts per token, and evaluate its training throughput. 

To realize this design, we propose Prototype-Mediated Routing (PMR). In the first stage, short-term states summarize preceding tokens and are combined with learnable prototypes to form adapted prototypes for the current sequence. PMR matches each token against these adapted prototypes to obtain a prototype distribution that captures its relative matching strengths. In the second stage, a prototype–expert association table defines a conditional distribution over experts for each prototype, allowing different prototypes to share experts and each prototype to favor multiple experts. Congestion prices derived from observed dispatch adjust the association scores that define these distributions, providing load feedback without an auxiliary load-balancing loss. For fixed token representations and matching states, this adjustment leaves the current prototype distribution unchanged. Finally, PMR mixes the conditional expert distributions using the token's responses as weights and selects the experts with the highest resulting probabilities.

Our main contributions are:
\begin{itemize}
    \item We propose PMR, a two-stage MoE router that separately parameterizes input matching and expert allocation. It combines token distributions over shared prototypes with prototype-conditioned expert distributions, explicitly representing many-to-many prototype--expert associations.

    \item We introduce separate mechanisms for context adaptation and load regulation. Short-term states adapt prototypes using preceding tokens, while congestion prices derived from dispatch feedback adjust prototype--expert association scores without an auxiliary load-balancing loss.

    \item Across three pretraining configurations, PMR improves seven-benchmark mean zero-shot accuracy by 1.45--2.65 percentage points over linear routers. Across five 7B--8B backbones, it improves four-task mean fine-tuning accuracy by 0.05--1.76 percentage points over MixLoRA.

    \item We develop a causal two-pass approximation for parallel blockwise state processing. On PMR-V3, it achieves $21.82\times$ the training throughput of eager tokenwise PMR, with 99.73\% exact Top-2 set agreement under teacher-forced evaluation.
\end{itemize}

\section{Preliminaries and Related Work}
\noindent\textbf{Mixture-of-Experts routing.}
A sparse Mixture-of-Experts (MoE) layer replaces a dense feed-forward network with $E$ expert networks $\{f_j\}_{j=1}^{E}$ and a router that determines which experts process each input token~\cite{shazeer2017outrageously,fedus2022switch}. Given a token representation $h_t\in\mathbb{R}^{d}$ at the current MoE layer, the conventional linear--softmax router first computes a distribution over the experts:

\begin{equation}
g_t=\operatorname{softmax}(U^\top h_t),
\label{eq:linear-router}
\end{equation}
where $U=[u_1,\ldots,u_E]\in\mathbb{R}^{d\times E}$ contains one routing vector per expert, and the softmax normalizes the scores across all $E$ experts. Based on this distribution, the router selects the experts with the highest scores, denoted by their index set $\mathcal D_t$. The selected experts then process the token, and their outputs are aggregated according to the corresponding gate values:
\begin{equation}
\operatorname{MoE}(h_t)
=
\sum_{j\in \mathcal{D}_t}
\frac{g_{t,j}}{\sum_{\ell\in \mathcal{D}_t}g_{t,\ell}}
f_j(h_t).
\label{eq:moe-output}
\end{equation}
Thus, the routing distribution serves two closely related purposes: determining the experts activated for each token through $\mathcal D_t$ and weighting their contributions to the final output through the normalized gate values.

\noindent\textbf{Load balancing.}
Auxiliary balancing losses encourage more even expert utilization~\cite{shazeer2017outrageously,lepikhin2021gshard}, whereas Loss-Free Balancing adjusts expert selection through load-dependent routing biases without introducing an auxiliary loss~\cite{wang2024auxiliary}. PMR instead uses dispatch feedback to adjust the association scores between prototypes and experts.

\noindent\textbf{Routing representations and expert specialization.}
To improve routing representations, X-MoE computes token--expert scores on a low-dimensional hypersphere to alleviate representation collapse~\cite{chi2022representation}, while STAR combines task-supervised routing logits with structure-aware logits derived from an evolving principal subspace, using a learnable mixing matrix to form expert-specific routing vectors~\cite{park2026star}. L2R assigns multiple anchors to each expert and aggregates their matching scores into an expert logit~\cite{l2r}. RouterInterp examines how experts are used in trained models and finds that an expert can be associated with several distinct input features and that individual features can be shared across experts~\cite{lasy2026routerinterp}. PMR allows such sharing by mixing the expert distributions of prototypes according to each token's prototype distribution.

\section{Method}
\label{sec:PMR-method}
\begin{figure*}[t]
    \centering
    \includegraphics[width=\textwidth]{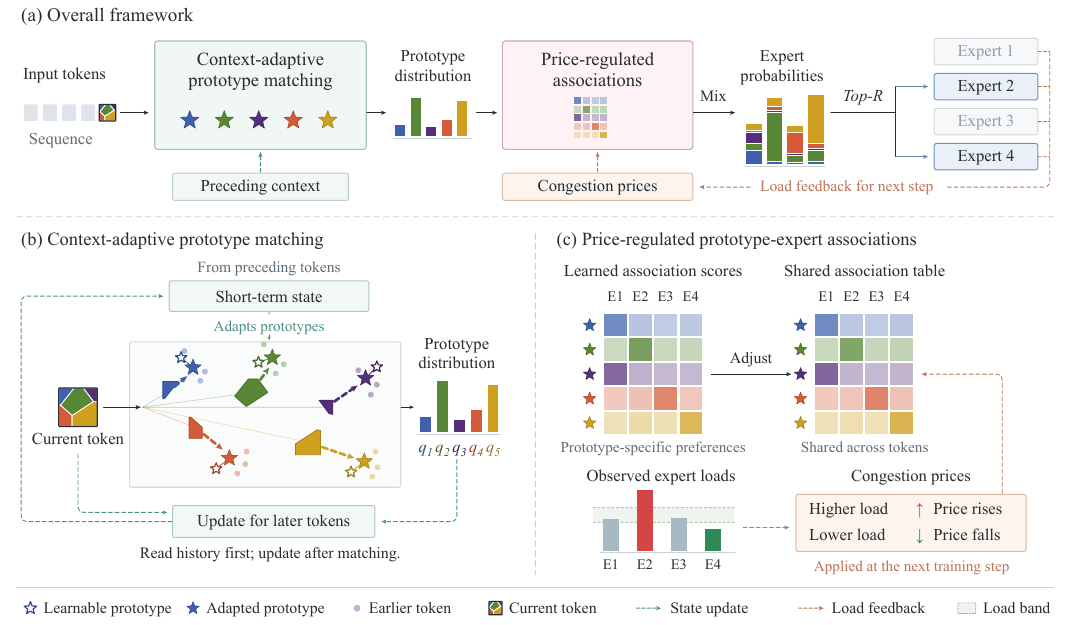}
    \caption{Overview of PMR.
    Context-adaptive prototype matching produces the response
    distribution $q_t$, which is combined with the shared
    prototype--expert association table $B$ to obtain expert
    probabilities $\pi_t$ for sparse dispatch.
    Short-term states adapt the matching references, while
    congestion prices regulate the association table using
    observed expert loads.}
    \label{fig:PMR-framework}
\end{figure*}
\subsection{Problem Formulation}
Given a token representation and its preceding sequence context, we seek to compute an expert distribution for Top-$R$ dispatch. To separate input matching from expert assignment, we model the token's distribution over prototypes and the conditional expert distribution associated with each feature prototype.

Let $Z \in \{1,\ldots,K\}$ index prototypes, $J \in \{1,\ldots,E\}$ index experts, and $h_{<t}$ denote the preceding sequence context. The prototype distribution $q_t\in\mathbb{R}^K$, with $q_{k,t}=p(Z=k\mid h_t,h_{<t})$, measures how strongly the current token responds to each prototype. PMR then maps these responses to expert probabilities through a prototype–expert association table $B\in\mathbb{R}^{K\times E}$ ($B_{kj}=p(J=j\mid Z=k)$), which is shared across tokens within each MoE layer for fixed model parameters and congestion prices. 

Under this model, its entries do not additionally depend on the current token or sequence context. The law of total probability gives
\begin{equation}
\pi_{j,t}
=
p(J=j \mid h_t, h_{<t})
=
\sum_{k=1}^{K} B_{kj}q_{k,t}.
\end{equation}

\subsection{Framework Overview}
 As illustrated in \cref{fig:PMR-framework}, PMR first uses Context-Adaptive Prototype Matching to compute a prototype distribution \(q_t\) for each token. The module projects and normalizes the input token representation \(h_t\) to \(z_t\) and matches \(z_t\) against the adapted prototype matrix \(M_{t-1}\), which combines learnable prototypes with the preceding sequence state. The resulting prototype distribution is passed to Price-Regulated Prototype–Expert Associations. This module applies the current congestion prices \(\lambda_n\) to the learned association scores \(C\) to construct a shared table \(B\) of conditional expert distributions. Mixing these distributions using \(q_t\) yields the expert probabilities \(\pi_t\) for Top-\(R\) dispatch. During training, the resulting dispatch counts are used to update the prices for the next step.

\subsection{Context-Adaptive Prototype Matching}
\label{sec:PMR-matching}

We project each input token representation $h_t\in\mathbb R^d$ into a lower-dimensional routing space ($d_p\ll d$) using a learned matrix $P\in\mathbb R^{d_p\times d}$: $z_t=\mathcal N_{\epsilon_z}(Ph_t)$, where $\mathcal N_\epsilon(v)=v/\max\{\|v\|_2,\epsilon\}$ for $\epsilon>0$ so that matching primarily reflects direction. While X-MoE~\cite{chi2022representation} also performs routing in a low-dimensional normalized space, PMR uses this projected space for token--prototype matching and learns $P$ through the task loss. Reducing the dimension lowers prototype-matching and state-update costs from $O(Kd)$ to $O(Kd_p)$ per token—at an additional $O(dd_p)$ projection cost—while reducing prototype and state storage.

In this projected space, PMR forms matching references shared across experts by combining learnable prototypes $A=[a_1,\ldots,a_K]\in\mathbb{R}^{d_p\times K}$ with short-term states $s_{k,t-1}$ that summarize preceding tokens. Each prototype is normalized after every optimizer update so that $\|a_k\|_2=1$ and remains fixed until the next update, whereas the short-term states evolve along the sequence. At position $t$, PMR combines each prototype with its preceding state:
\begin{equation}
\begin{aligned}
\tilde m_{k,t-1}
&=(1-\beta_{k,t-1})a_k+\beta_{k,t-1}s_{k,t-1},\\
m_{k,t-1}&=\mathcal N_{\epsilon_m}(\tilde m_{k,t-1}).
\end{aligned}
\label{eq:PMR-effective-prototype}
\end{equation}
Here, the mixing weight satisfies $0\leq\beta_{k,t-1}\leq\beta_{\max}<1$ and increases with the accumulated prototype response. Together with $\|s_{k,t-1}\|_2\leq1$, the unit-norm prototype constraint bounds the pre-normalization displacement by $\|\tilde m_{k,t-1}-a_k\|_2\leq2\beta_{k,t-1}\leq2\beta_{\max}$. Stacking the adapted prototypes into the $M_{t-1}=[m_{1,t-1},\ldots,m_{K,t-1}]$, we compute the current prototype distribution as
\begin{equation}
q_t=\operatorname{softmax}
\left(M_{t-1}^{\top}z_t/\tau_p\right),
\label{eq:PMR-response}
\end{equation}
where $\tau_p>0$ controls distribution concentration and $q_t$ describes the token's response strengths across prototypes.

After matching, PMR updates each short-term state
for subsequent tokens:
\begin{equation}
\begin{aligned}
\tilde{s}_{k,t}
&=s_{k,t-1}+\alpha_S
\bigl[q_{k,t}(z_t-s_{k,t-1})\\
&\qquad+\lambda_{sa}(a_k-s_{k,t-1})\bigr],\\
s_{k,t}
&=\operatorname{Proj}_{\mathbb{B}}(\tilde{s}_{k,t}).
\end{aligned}
\label{eq:PMR-state-update}
\end{equation}
Here, $\alpha_S>0$ is the adaptation step size, $\lambda_{sa}\geq0$ controls attraction toward the prototype, and $\operatorname{Proj}_{\mathbb B}(v)=v/\max\{1,\|v\|_2\}$ projects onto the unit ball. In this update, the response $q_{k,t}$ weights the current token's contribution, while the prototype term retains the long-term reference.

Accumulated prototype responses govern the influence of sequence history on subsequent matching:
\begin{equation}
\begin{aligned}
\nu_{k,t}&=\rho_\beta\nu_{k,t-1}
             +\operatorname{sg}(q_{k,t}),\\
\beta_{k,t}&=\beta_{\max}
\frac{\nu_{k,t}}{\nu_{k,t}+\kappa_\beta}.
\end{aligned}
\label{eq:PMR-mixing-update}
\end{equation}
Here, $\nu_{k,t}$  is the accumulated prototype response, $0\leq\rho_\beta<1$ controls its exponential decay, $\kappa_\beta>0$ sets the response scale, and $\operatorname{sg}$ denotes stop-gradient, so that stronger historical activation gives the short-term state greater influence up to $\beta_{\max}$.

At logical-sequence boundaries—including packed-sequence boundaries—short-term states and accumulated prototype responses reset to zero, while padding positions leave them unchanged; consequently, the first token uses the learnable prototypes alone as matching references. Within a sequence, reading preceding states and mixing weights before writing current information preserves causality. These reads are detached, as are the copies of $q_t$, $z_t$, and $a_k$ used in \cref{eq:PMR-state-update}, while the current matching computation retains its task-gradient path.

\noindent\textbf{Blockwise implementation.}
To reduce the sequential overhead of short-term state updates during training and prompt prefilling, we design a causal two-pass blockwise approximation to the tokenwise recurrence. By estimating the preceding states through prefix accumulation, this scheme enables parallel response computation within each block.

Consider a block of $L$ valid tokens, indexed by $i=1,\ldots,L$,
with incoming short-term states $S_0=[s_{1,0},\ldots,s_{K,0}]$ and accumulated
prototype responses $\nu_0$. Let $M(S,\nu)$ denote the adapted prototype matrix
constructed from these quantities using~\cref{eq:PMR-effective-prototype}
and~\cref{eq:PMR-mixing-update}. The first pass predicts all
responses in parallel using the block-entry states:
\begin{equation}
q_i^{(0)}=\operatorname{softmax}\!\left(
M(S_0,\nu_0)^\top z_i/\tau_p\right).
\label{eq:PMR-block-predict}
\end{equation}
For pass $r\in\{0,1\}$, define the update direction at the
block-entry state as
\begin{equation}
\begin{aligned}
\xi_{k,j}^{(r)}
={}&q_{k,j}^{(r)}(z_j-s_{k,0}) + \lambda_{sa}(a_k-s_{k,0}).
\end{aligned}
\label{eq:PMR-block-direction}
\end{equation}
Exclusive prefix sums of the predicted updates provide estimates
of the states preceding each position:
\begin{equation}
\begin{aligned}
\widehat s_{k,i-1}
&=\operatorname{Proj}_{\mathbb B}\!\left(
s_{k,0}+\alpha_S\sum_{j=1}^{i-1}\xi_{k,j}^{(0)}\right),\\
\widehat\nu_{k,i-1}
&=\rho_\beta^{i-1}\nu_{k,0}
+\sum_{j=1}^{i-1}\rho_\beta^{i-1-j}q_{k,j}^{(0)}.
\end{aligned}
\label{eq:PMR-block-prefix}
\end{equation}
For the predicted responses, the expression for $\widehat\nu$
exactly unrolls the decay recurrence, while $\widehat s$
approximates the state recurrence using update directions
evaluated at $s_{k,0}$. The second pass uses these estimates
to correct the responses in parallel:
\begin{equation}
q_i^{(1)}=\operatorname{softmax}\!\left(
M(\widehat S_{i-1},\widehat\nu_{i-1})^\top z_i/\tau_p\right),
\label{eq:PMR-block-correct}
\end{equation}
where
$\widehat S_{i-1}=[\widehat s_{1,i-1},\ldots,\widehat s_{K,i-1}]$.
The corrected prototype distributions $q_i^{(1)}$ are used for expert allocation
and to compute the outgoing states:
\begin{equation}
\begin{aligned}
s_k^{\mathrm{out}}
&=\operatorname{Proj}_{\mathbb B}\!\left(
s_{k,0}+\alpha_S\sum_{i=1}^{L}\xi_{k,i}^{(1)}\right),\\
\nu_k^{\mathrm{out}}
&=\rho_\beta^L\nu_{k,0}
+\sum_{i=1}^{L}\rho_\beta^{L-i}q_{k,i}^{(1)}.
\end{aligned}
\label{eq:PMR-block-carry}
\end{equation}
The block-end state update also uses $s_{k,0}$ as its reference.
We carry $(S^{\mathrm{out}},\nu^{\mathrm{out}})$ into the next block,
following the same stop-gradient, sequence-boundary, and padding
conventions as above. Exclusive prefixes preserve causality, while
the two parallel passes and prefix accumulation avoid sequential
tokenwise state updates within each block. This enables parallel state processing during training and the prompt-prefilling stage of inference while retaining history across blocks. Under this blockwise scheme, $q_t$ and $M_{t-1}$ in the subsequent allocation and analysis denote the corrected prototype distribution $q_i^{(1)}$ and the estimated adapted prototype matrix $M(\widehat S_{i-1},\widehat\nu_{i-1})$ at the corresponding block position $i$.
Algorithm~\ref{alg:PMR-blockwise} summarizes the two-pass
computation for one block.

\par\smallskip
\begin{figure}[!t]
\centering
\noindent\begin{minipage}{\columnwidth}
\small
\refstepcounter{PMRalgorithm}\label{alg:PMR-blockwise}

\hrule
\vspace{3pt}
\noindent\textbf{Algorithm \thePMRalgorithm.}
Causal Two-Pass Blockwise Prototype Matching
\par\vspace{3pt}
\hrule
\vspace{4pt}

\noindent\textbf{Input:}
Normalized projected token representations $\{z_i\}_{i=1}^{L}$,
prototype matrix \(A\), and incoming state $(S_0,\nu_0)$.

\noindent\textbf{Output:}
Corrected prototype distributions $\{q_i^{(1)}\}_{i=1}^{L}$ and
outgoing state $(S^{\mathrm{out}},\nu^{\mathrm{out}})$.

\begin{enumerate}
\setlength{\itemsep}{1pt}
\setlength{\parsep}{0pt}
\item Form the block-entry adapted prototype matrix $M(S_0,\nu_0)$ from
      Eqs.~\eqref{eq:PMR-effective-prototype}
      and~\eqref{eq:PMR-mixing-update}.
\item Compute the first-pass prototype distributions $q_i^{(0)}$ in parallel
      via Eq.~\eqref{eq:PMR-block-predict}.
\item Estimate $(\widehat S_{i-1},\widehat\nu_{i-1})$ from
      first-pass responses at positions $j<i$ via
      Eqs.~\eqref{eq:PMR-block-direction}
      and~\eqref{eq:PMR-block-prefix}.
\item Compute the corrected prototype distributions $q_i^{(1)}$ in parallel
      from the estimated states via Eq.~\eqref{eq:PMR-block-correct}.
\item Compute $(S^{\mathrm{out}},\nu^{\mathrm{out}})$ from
      the corrected responses via
      Eqs.~\eqref{eq:PMR-block-direction}
      and~\eqref{eq:PMR-block-carry}.
\end{enumerate}

\hrule
\end{minipage}
\end{figure}
\par\smallskip

\subsection{Price-Regulated Prototype--Expert Associations}
\label{sec:PMR-assignment}

Given the prototype distribution $q_t$, this stage constructs a shared prototype–expert association table $B$—whose rows specify prototype-conditioned expert distributions—to translate prototype responses into expert probabilities. Let $C\in\mathbb{R}^{K\times E}$ contain learned association scores, where a larger $C_{kj}$ indicates a stronger preference for assigning inputs matching prototype $k$ to expert $j$. Task training shapes these preferences, while congestion prices adjust allocations based on actual load. Whereas Loss-Free Balancing~\cite{wang2024auxiliary} applies load-dependent biases directly to expert scores for selection, PMR incorporates load feedback into the prototype--expert association scores before applying the row-wise softmax to construct $B$.

The prices used at training step $n$ depend on completed earlier steps. For step $n-1$ with $T_{n-1}>0$ valid tokens, let $\mathcal D_{n-1,i}$ contain the $R$ distinct experts executed for token $i$ ($1\leq R\leq E$). The observed dispatch share is $f_{n-1,j}=(RT_{n-1})^{-1}\sum_i\mathbf1\{j\in\mathcal D_{n-1,i}\}$, with counts aggregated across micro-batches and devices before normalization. Given the tolerance interval $[\ell,u]=[(1-\delta)/E,(1+\delta)/E]$ for $\delta\geq0$, we define the signed residual $c_{n-1,j}=f_{n-1,j}-\operatorname{clip}(f_{n-1,j},\ell,u)$, where positive and negative residuals identify utilization above and below the interval around $1/E$. Congestion prices accumulate these residuals according to
\begin{equation}
\lambda_n=\operatorname{Center}
\left(\lambda_{n-1}+\eta_\lambda c_{n-1}\right),
\label{eq:PMR-price-update}
\end{equation}
where $\eta_\lambda>0$ is the correction step size and $\operatorname{Center}$ subtracts the mean across experts.

Prices are initialized to zero: a valid step's price update is committed after a successful optimizer update for use in the next step (while empty or failed steps leave prices unchanged), so step $n$ operates with a fixed congestion price vector $\lambda_n\in\mathbb R^E$. 
The price-adjusted association scores are obtained by subtracting each expert’s congestion price uniformly across prototypes:
\begin{equation}
\begin{aligned}
\widehat C=C-\boldsymbol{1}_K\lambda_{n}^{T},\quad
\widehat C_{kj}=C_{kj}-\lambda_{n,j}.
\end{aligned}
\label{eq:PMR-price-adjusted-scores}
\end{equation}
Here, \(\mathbf{1}_K \in \mathbb{R}^K\) is the all-ones column vector. The learned association score \(C_{kj}\) expresses a task-dependent association preference, and \(\lambda_{n,j}\) is expert \(j\)'s congestion price. The same congestion price is subtracted from every entry in the corresponding expert column.

We construct $B$ by balancing the adjusted association preferences against an entropy regularizer:
\begin{equation}
B=\arg\max_{\substack{X\geq0\\X\mathbf1_E=\mathbf1_K}}
\left\{
\sum_{k,j}X_{kj}\widehat C_{kj}
+\tau_e\sum_k H(X_{k,:})
\right\},
\label{eq:PMR-allocation}
\end{equation}
where $H(p)=-\sum_jp_j\log p_j$ ($0\log0=0$) and $\tau_e>0$. In this objective, the adjusted scores favor particular experts while entropy favors more diffuse conditional expert distributions—with $\tau_e$ balancing this trade-off—and the row constraints normalize the resulting conditional expert distributions.

When all entries of \(\widehat C\) are finite real numbers, the objective is strictly concave in each row over the probability simplex, and the entropy–softmax relation therefore yields the unique closed-form solution:
\begin{equation}
B=\operatorname{softmax}_{\mathrm{row}}
\left(\widehat C/\tau_e\right).
\label{eq:PMR-association}
\end{equation}
For fixed token representations, prototype parameters, and short-term states, price feedback adjusts $B$ without changing $q_t$, allowing allocation to respond to load while preserving the current prototype responses. Because the allocation objective defines $B$ through its closed-form solution rather than being added to the task loss, PMR is trained directly with task cross-entropy while price feedback regulates load without an auxiliary balancing loss.

\noindent\textbf{Expert probabilities and properties.}
Finally, we apply the association table $B$ to each token's prototype distribution $q_t$—or in batch to prototype distributions $Q=[q_1,\ldots,q_T]$ collected within the same MoE layer and training step—to obtain the expert probabilities:
\begin{equation}
\pi_t=B^\top q_t,\quad \Pi=B^\top Q.
\label{eq:PMR-dispatch}
\end{equation}
Because $q_t$ and each row of $B$ sum to one, $\pi_t$ forms a valid probability distribution in which each product $q_{k,t}B_{kj}$ records prototype $k$'s contribution to expert $j$, allowing token--prototype responses and prototype--expert associations to be inspected separately. 

For sparse execution, PMR selects the $R$ experts with the highest probabilities in $\pi_t$ and renormalizes the corresponding entries to combine their outputs, as in ~\cref{eq:moe-output}.

For a fixed adapted prototype matrix $M_{t-1}$ and association
table $B$, the probability mixture yields the following log-probability
ratio between experts $j$ and $\ell$:
\begin{equation}
\begin{aligned}
\log\frac{\pi_{j,t}}{\pi_{\ell,t}}
={}&
\log\sum_{k=1}^{K} B_{kj}
e^{m_{k,t-1}^{\top}z_t/\tau_p}\\
&-
\log\sum_{k=1}^{K} B_{k\ell}
e^{m_{k,t-1}^{\top}z_t/\tau_p}.
\end{aligned}
\label{eq:PMR-relative-preference}
\end{equation}
Whereas the corresponding log-probability ratio is affine in $z_t$
for linear--softmax routing on the same representation,
combining multiple prototype responses at the probability level in PMR
allows this ratio to vary nonlinearly with $z_t$.

At the prototype-distribution level, mixing through a fixed
$B$ is also non-expansive in total variation: for any two
prototype distributions $q$ and $q'$,
\begin{equation}
\operatorname{TV}(B^\top q,B^\top q')
\leq \operatorname{TV}(q,q'),
\label{eq:PMR-nonexpansiveness}
\end{equation}
where $\operatorname{TV}(p,p')=\frac{1}{2}\|p-p'\|_1$.
This bound follows from the triangle inequality—since $B$ has
nonnegative entries and each row sums to one—ensuring that
similar prototype distributions yield similar expert
probability distributions. This bound concerns the probabilities before Top-\(R\) selection.

\section{Experiments}
In this section, we evaluate PMR in MoE language model pretraining and parameter-efficient fine-tuning. We further assess its key components and analyze routing behavior, expert contributions, and computational efficiency.

\subsection{Experimental Setup for Pretraining}
\label{sec:experimental-setup}

\subsubsection{Data}
The pretraining corpus comprises FineWeb-Edu~\cite{penedo2024fineweb} and Cosmopedia v2~\cite{benallal2024smollmcorpus}. FineWeb-Edu provides web text filtered for educational quality, while Cosmopedia v2 consists of synthetic textbooks, blog posts, and stories. The corpus is divided into two non-overlapping stages, each containing 4B tokens, with the two datasets mixed at an approximate 89:11 token ratio in each stage. We conduct zero-shot evaluation using the Language Model Evaluation Harness (v0.4.12)~\cite{gao2024harness} on seven benchmarks covering diverse language understanding and reasoning capabilities. Specifically, HellaSwag~\cite{zellers2019hellaswag} evaluates commonsense reasoning through scenario continuation, PIQA~\cite{bisk2020piqa} assesses physical commonsense reasoning and practical problem solving, and WinoGrande~\cite{sakaguchi2020winogrande} tests commonsense reasoning through fill-in-the-blank disambiguation. ARC-Easy and ARC-Challenge~\cite{clark2018arc} evaluate grade-school science question answering at different difficulty levels, while OpenBookQA~\cite{mihaylov2018openbookqa} focuses on scientific question answering that combines factual knowledge with reasoning. Finally, LAMBADA~\cite{paperno2016lambada} evaluates context-dependent language modeling through final-word prediction. The dataset sizes of all seven benchmarks are reported in \cref{tab:benchmark_sizes}.

\begin{table}[t!]
\caption{Dataset sizes of the seven zero-shot evaluation benchmarks.}
\label{tab:benchmark_sizes}
\centering
\footnotesize
\setlength{\tabcolsep}{8pt}
\begin{tabular}{lc}
\hline
Benchmark & Dataset Size (\# Samples) \\
\hline
HellaSwag & 59,950 \\
PIQA & 21,035 \\
WinoGrande-XL & 43,432 \\
ARC-Easy & 5,197 \\
ARC-Challenge & 2,590 \\
OpenBookQA & 5,957 \\
LAMBADA & 10,022 \\
\hline
\end{tabular}
\end{table}

\subsubsection{Implementation Details}
We conduct pretraining using three Mixtral configurations~\cite{jiang2024mixtral}, namely TinyMixtral-V1, TinyMixtral-V2, and TinyMixtral-V3. Their counterparts with PMR routing are TinyMixtral-V1 + PMR (PMR-V1), TinyMixtral-V2 + PMR (PMR-V2), and TinyMixtral-V3 + PMR (PMR-V3), respectively. TinyMixtral-V1 provides an initial small-scale setting for developing PMR, TinyMixtral-V2 extends the model scale, while TinyMixtral-V3 adopts a more parameter-efficient architecture that achieves performance comparable to TinyMixtral-V2 with fewer parameters. Their baseline architectures are summarized in \cref{tab:configs-TM}, where parameter counts refer to the baseline models before replacing the router with PMR. Active counts include shared parameters and the two experts selected per token in each layer. All three configurations use grouped-query attention~\cite{ainslie2023gqa}, SwiGLU feed-forward experts~\cite{shazeer2020glu}, RMSNorm~\cite{zhang2019rmsnorm}, a 32{,}000-token vocabulary, tied input and output embeddings, and Top-2 expert execution. PMR replaces TinyMixtral's linear--softmax router and auxiliary load-balancing loss~\cite{lepikhin2021gshard,fedus2022switch} with its two-stage routing architecture and congestion-pricing mechanism, and the key PMR routing configurations are reported in \cref{tab:configs-TM}. All three PMR configurations apply blockwise state processing.

We train the models with an autoregressive next-token prediction objective using AdamW~\cite{loshchilov2019adamw} and a sequence length of 1{,}024 tokens. The learning rate increases linearly to $7\times10^{-4}$ over the first 2{,}000 optimizer updates, followed by cosine decay over the remaining training budget. TinyMixtral-V1 and PMR-V1 are trained on 4B tokens, while the V2 and V3 baseline-PMR pairs are trained on 8B tokens, with the learning-rate schedule continuing across the two data stages without restarting. Zero-shot performance is evaluated using the Language Model Evaluation Harness~\cite{gao2024harness} on the aforementioned seven benchmarks. We report individual benchmark accuracies and their unweighted average, with length normalization enabled when scoring for HellaSwag, ARC-Challenge, and OpenBookQA.

\begin{table}[!t]
\caption{Architectural configurations of the three TinyMixtral models and key PMR routing configurations with their paired TinyMixtral architectures in pretraining experiments.}
\label{tab:configs-TM}
\centering
\footnotesize
\setlength{\tabcolsep}{1.25pt}
\begin{tabular*}{\columnwidth}{@{\extracolsep{\fill}}>{\raggedright\arraybackslash}p{0.47\columnwidth}rrr@{}}
\hline
\noalign{\vskip 3pt}
Configuration & V1 & V2 & V3 \\
\hline
Total parameters (M) & 432.39 & 1182.17 & 477.47 \\
Active parameters (M) & 175.53 & 351.70 & 276.14 \\
Layers & 10 & 16 & 16 \\
Hidden dimensionality & 896 & 1024 & 1024 \\
Q/KV heads & 14 / 2 & 16 / 4 & 16 / 4 \\
Expert dimensionality & 2389 & 2816 & 2048 \\
Experts per layer & 6 & 8 & 4 \\
Query/key normalization & No & No & Yes \\
Training tokens & 4B & 8B & 8B \\
\hline
PMR routing & PMR-V1 & PMR-V2 & PMR-V3 \\
\hline
Prototypes ($K$) & 12 & 16 & 8 \\
Projection dimension ($d_p$) & 128 & 15 & 7 \\
Block size (tokens) & 128 & 128 & 16 \\
\hline
\end{tabular*}
\end{table}


\begin{table*}[!t]
\caption{Zero-shot evaluation accuracy (\%) of three TinyMixtral configurations and their PMR-enhanced counterparts on seven benchmarks. Avg. denotes the unweighted mean accuracy across the seven benchmarks. Best results are in bold.}
\label{tab:PMR_pretraining_results}
\centering
\begingroup
\small
\setlength{\tabcolsep}{1.5pt}
\setlength{\fboxsep}{2pt}
\renewcommand{\arraystretch}{1.28}
\begin{tabular*}{\textwidth}{@{\extracolsep{\fill}}l*{10}{c}@{}}
\noalign{\hrule height 0.8pt}
\rule{0pt}{2.8ex}Method & \shortstack{Total\\Params} & \shortstack{Active\\Params}
& HellaS. & PIQA & WinoG. & ARC-E & ARC-C & OBQA
& LAMBADA & Avg. \\[0.65ex]
\hline
\multicolumn{11}{@{}l@{}}{\makebox[0pt][l]{%
\colorbox[gray]{0.95}{%
\makebox[\dimexpr\textwidth-2\fboxsep\relax][c]{%
\rule[-0.3ex]{0pt}{2.1ex}\textit{4B token pre-training (TinyMixtral-V1)}}}}} \\[0.3ex]
TinyMixtral-V1 & 432.39M & 175.53M
& 31.00 & 61.30 & 50.80 & 42.20 & 24.70 & 30.80 & 24.00 & 37.83 \\
TinyMixtral-V1 + PMR & 433.50M & 176.64M
& \textbf{31.57} & \textbf{61.70} & \textbf{51.62}
& \textbf{48.91} & \textbf{25.43} & \textbf{31.80}
& \textbf{24.57} & \textbf{39.37} \\[0.35ex]
\hline
\multicolumn{11}{@{}l@{}}{\makebox[0pt][l]{%
\colorbox[gray]{0.95}{%
\makebox[\dimexpr\textwidth-2\fboxsep\relax][c]{%
\rule[-0.3ex]{0pt}{2.1ex}\textit{8B token pre-training (TinyMixtral-V2)}}}}} \\[0.3ex]
TinyMixtral-V2 & 1182.17M & 351.70M
& 32.90 & 63.00 & 52.30 & 47.90 & \textbf{27.90}
& 30.60 & 23.40 & 39.71 \\
TinyMixtral-V2 + PMR & 1182.31M & 351.84M
& \textbf{37.25} & \textbf{64.69} & \textbf{52.41} & \textbf{53.70}
& 26.96 & \textbf{31.20} & \textbf{30.31} & \textbf{42.36} \\[0.35ex]
\hline
\multicolumn{11}{@{}l@{}}{\makebox[0pt][l]{%
\colorbox[gray]{0.95}{%
\makebox[\dimexpr\textwidth-2\fboxsep\relax][c]{%
\rule[-0.3ex]{0pt}{2.1ex}\textit{8B token pre-training (TinyMixtral-V3)}}}}} \\[0.3ex]
TinyMixtral-V3 & 477.47M & 276.14M
& 33.61 & 63.55 & \textbf{51.54} & 51.09 & \textbf{26.37}
& 29.20 & 25.25 & 40.09 \\
TinyMixtral-V3 + PMR & 477.52M & 276.19M
& \textbf{35.43} & \textbf{63.82} & 49.09 & \textbf{55.01}
& 25.34 & \textbf{33.60} & \textbf{28.47} & \textbf{41.54} \\[0.35ex]
\noalign{\hrule height 0.8pt}
\end{tabular*}
\endgroup
\end{table*}

\subsection{Experimental Setup for Parameter-Efficient Fine-Tuning}

\subsubsection{Data}
We use the official training splits of PIQA~\cite{bisk2020piqa}, ARC-Easy and ARC-Challenge~\cite{clark2018arc}, and OpenBookQA~\cite{mihaylov2018openbookqa} for task-specific fine-tuning. We construct a prompt containing the question or goal and the answer choices, each marked with an option identifier such as A or B. For training, we append the identifier of the correct option to the end of the prompt to form the training sequence. We evaluate on the PIQA validation split and the test splits of ARC-Easy, ARC-Challenge, and OpenBookQA. For evaluation, we use the same prompt format without appending the answer identifier. Fine-tuning and evaluation are performed separately for each task. We report accuracy for all four tasks.

\subsubsection{Implementation Details}
We evaluate five officially released pretrained dense backbones from the Llama and Qwen families, covering both established and more recent model generations. Their moderate parameter sizes make them suitable for parameter-efficient fine-tuning, while their dense architectures provide a consistent starting point for constructing MoE models. The selected backbones include Llama-2-7B~\cite{touvron2023llama2}, Llama-3-8B and Llama-3.1-8B-Base~\cite{grattafiori2024llama3}, Qwen2.5-7B~\cite{qwen2025qwen25}, and Qwen3-8B-Base~\cite{yang2025qwen3}. Following MixLoRA~\cite{li2024mixlora}, we place eight low-rank adaptation (LoRA) experts~\cite{hu2022lora} in the feed-forward layer of each Transformer block and activate two experts per token. Each expert comprises LoRA updates to the FFN up, down, and gate projections. We additionally apply LoRA to the query, key, value, and output projections in the attention layers. All LoRA modules use rank $r=16$ and scaling parameter $\alpha=32$. We compare MixLoRA with MixLoRA + PMR under the same expert architecture and Top-2 execution, where PMR replaces the MixLoRA router with its two-stage routing architecture and congestion-pricing mechanism in place of the auxiliary routing losses. The backbone parameters remain frozen, while only the LoRA and learnable routing parameters are optimized.


We fine-tune the models with an autoregressive next-token prediction objective using AdamW with a constant learning rate of $2\times10^{-4}$ and no warmup. Training runs for two epochs with a maximum sequence length of 512 and an effective batch size of 16. During training, next-token cross-entropy is computed over the full prompt-and-answer sequence. PMR uses $K=16$ prototypes, a projection dimension of $d_p=15$, and a block size of 16 tokens.
Note that congestion prices are updated from observed expert dispatch during fine-tuning and held fixed during evaluation, and blockwise state processing is applied. For evaluation on the aforementioned four benchmarks, the model predicts by selecting the candidate option with the highest prediction logit, where we report individual benchmark accuracies and their unweighted average.

\begin{table}[!t]
\caption{Fine-tuning accuracy (\%) on four reasoning benchmarks across five pretrained backbones. Base represents the unadapted backbone. Best results are in bold.}
\label{tab:PMR_finetuning_results}
\centering
\begingroup
\footnotesize
\setlength{\tabcolsep}{0.5pt}
\setlength{\fboxsep}{2pt}
\renewcommand{\arraystretch}{0.9}
\begin{tabular*}{\columnwidth}{@{\extracolsep{\fill}}l*{5}{c}@{}}
\noalign{\hrule height 0.8pt}
\rule{0pt}{2.8ex}Method
& ARC-C
& ARC-E
& OBQA
& PIQA
& Avg. \\[0.5ex]
\hline

\multicolumn{6}{@{}l@{}}{\makebox[0pt][l]{%
\colorbox[gray]{0.95}{%
\makebox[\dimexpr\columnwidth-2\fboxsep\relax][c]{%
\rule[-0.3ex]{0pt}{2.1ex}\textit{Llama-2-7B}}}}} \\
Base
& 32.51 & 40.24 & 34.00 & 56.91 & 40.91 \\
MixLoRA
& 54.78 & 77.86 & 77.20 & \textbf{81.77} & 72.90 \\
MixLoRA + PMR
& \textbf{59.39} & \textbf{77.99} & \textbf{79.80}
& 81.45 & \textbf{74.66} \\
\hline

\multicolumn{6}{@{}l@{}}{\makebox[0pt][l]{%
\colorbox[gray]{0.95}{%
\makebox[\dimexpr\columnwidth-2\fboxsep\relax][c]{%
\rule[-0.3ex]{0pt}{2.1ex}\textit{Llama-3-8B}}}}} \\
Base
& 66.30 & 81.31 & 67.80 & 61.86 & 69.32 \\
MixLoRA
& 76.28 & 87.33 & 83.80 & \textbf{85.91} & 83.33 \\
MixLoRA + PMR
& \textbf{76.96} & \textbf{89.06} & \textbf{85.40}
& 85.64 & \textbf{84.26} \\
\hline

\multicolumn{6}{@{}l@{}}{\makebox[0pt][l]{%
\colorbox[gray]{0.95}{%
\makebox[\dimexpr\columnwidth-2\fboxsep\relax][c]{%
\rule[-0.3ex]{0pt}{2.1ex}\textit{Llama-3.1-8B-Base}}}}} \\
Base
& 68.09 & 81.10 & 69.20 & 66.70 & 71.27 \\
MixLoRA
& \textbf{78.16} & 87.16 & 85.00 & 85.91 & 84.06 \\
MixLoRA + PMR
& 77.05 & \textbf{88.43} & \textbf{86.20}
& \textbf{86.83} & \textbf{84.63} \\
\hline

\multicolumn{6}{@{}l@{}}{\makebox[0pt][l]{%
\colorbox[gray]{0.95}{%
\makebox[\dimexpr\columnwidth-2\fboxsep\relax][c]{%
\rule[-0.3ex]{0pt}{2.1ex}\textit{Qwen2.5-7B}}}}} \\
Base
& 85.84 & 91.41 & 84.40 & 82.43 & 86.02 \\
MixLoRA
& 86.43 & \textbf{92.51} & \textbf{93.00}
& 88.14 & 90.02 \\
MixLoRA + PMR
& \textbf{87.03} & 92.05 & 92.80
& \textbf{88.41} & \textbf{90.07} \\
\hline

\multicolumn{6}{@{}l@{}}{\makebox[0pt][l]{%
\colorbox[gray]{0.95}{%
\makebox[\dimexpr\columnwidth-2\fboxsep\relax][c]{%
\rule[-0.3ex]{0pt}{2.1ex}\textit{Qwen3-8B-Base}}}}} \\
Base
& \textbf{90.02} & 93.31 & 85.20
& 86.18 & 88.68 \\
MixLoRA
& 88.82 & 92.68 & \textbf{93.20}
& 89.50 & 91.05 \\
MixLoRA + PMR
& 89.93 & \textbf{93.35} & \textbf{93.20}
& \textbf{89.72} & \textbf{91.55} \\
\noalign{\hrule height 0.8pt}
\end{tabular*}
\endgroup
\end{table}

\begin{table*}[!t]
\caption{Zero-shot evaluation accuracy (\%) on seven benchmarks for TinyMixtral-V1 trained on 4B tokens under different routing and load-balancing designs. Best results are in bold.}
\label{tab:PMR_routing_ablation}
\centering
\begingroup
\small
\setlength{\tabcolsep}{4pt}
\setlength{\fboxsep}{2pt}
\renewcommand{\arraystretch}{1.28}

\begin{tabular*}{\textwidth}{@{\extracolsep{\fill}}l*{8}{c}@{}}
\noalign{\hrule height 0.8pt}

\rule{0pt}{2.8ex}Configuration
& HellaS.
& PIQA
& WinoG.
& ARC-E
& ARC-C
& OBQA
& LAMBADA
& Avg. \\[0.65ex]

\hline

Linear + auxiliary
& 31.00 & 61.30 & 50.80 & 42.20
& 24.70 & 30.80 & 24.00 & 37.83 \\

PMR + auxiliary
& \textbf{31.67} & \textbf{62.62}
& 50.91 & 46.00
& 25.09 & 29.20
& \textbf{25.95} & 38.78 \\

PMR + prices
& 31.57 & 61.70
& \textbf{51.62} & \textbf{48.91}
& \textbf{25.43} & \textbf{31.80}
& 24.57 & \textbf{39.37} \\[0.35ex]

\noalign{\hrule height 0.8pt}
\end{tabular*}
\endgroup
\end{table*}

\subsection{Language Model Pretraining and Zero-Shot Evaluation}

\cref{tab:PMR_pretraining_results} reports zero-shot results on seven benchmarks for the three TinyMixtral configurations. PMR achieves average scores of 39.37, 42.36, and 41.54 for PMR-V1, PMR-V2, and PMR-V3, respectively, exceeding the corresponding baselines by 1.54, 2.65, and 1.45 percentage points. It improves performance on all seven tasks for PMR-V1, six of seven tasks for PMR-V2, and five of seven tasks for PMR-V3. ARC-Easy shows consistent gains across all three configurations, with improvements of 6.71, 5.80, and 3.92 percentage points, respectively. Other notable gains include HellaSwag and LAMBADA for PMR-V2, which improve by 4.35 and 6.91 points, respectively. In contrast, ARC-Challenge shows small changes in both directions, while WinoGrande improves slightly for PMR-V1 and PMR-V2 but declines by 2.45 percentage points for PMR-V3. These improvements incur only a small parameter overhead, with PMR increasing total parameters by at most 0.26\% and active parameters by less than 0.64\% relative to the corresponding baselines.

\subsection{Parameter-Efficient Fine-Tuning of Language Models}

Tabulated in \cref{tab:PMR_finetuning_results}, PMR improves the four-task mean over MixLoRA~\cite{li2024mixlora} across all five backbones, with gains of 1.76, 0.93, and 0.57 percentage points on Llama-2-7B, Llama-3-8B, and Llama-3.1-8B-Base, respectively, and 0.05 and 0.50 percentage points on Qwen2.5-7B and Qwen3-8B-Base. The improvements are also reflected at the task level. Each Llama backbone achieves higher accuracy on three of the four benchmarks, with the largest gains observed on ARC-Challenge (+4.61 pp) and OpenBookQA (+2.60 pp) for Llama-2-7B. As for Qwen3-8B-Base, PMR improves results on ARC-Challenge and ARC-Easy by 1.11 and 0.67 pp, respectively, while the gains on Qwen2.5-7B are more marginal. These results demonstrate that PMR consistently improves the overall fine-tuning performance across diverse backbones, while the magnitude and distribution of task-level gains vary across models. Combined with the pretraining results, they further demonstrate the effectiveness of PMR in both MoE pretraining and parameter-efficient adaptation of frozen dense backbones.

\begin{table}[!t]
\caption{Fine-tuning accuracy (\%) with and without short-term sequence state across four reasoning benchmarks and five pretrained backbones. Best results are in bold.}
\label{tab:PMR-state-ablation}
\centering
\begingroup
\footnotesize
\setlength{\tabcolsep}{0.5pt}
\setlength{\fboxsep}{2pt}
\renewcommand{\arraystretch}{1.25}
\begin{tabular*}{\columnwidth}{@{\extracolsep{\fill}}l*{5}{c}@{}}
\noalign{\hrule height 0.8pt}

\rule{0pt}{2.8ex}Method
& ARC-C
& ARC-E
& OBQA
& PIQA
& Avg. \\[0.5ex]

\hline

\multicolumn{6}{@{}l@{}}{\makebox[0pt][l]{%
\colorbox[gray]{0.95}{%
\makebox[\dimexpr\columnwidth-2\fboxsep\relax][c]{%
\rule[-0.3ex]{0pt}{2.1ex}\textit{Llama-2-7B}}}}} \\[0.3ex]

MixLoRA + PMR
& \textbf{59.39}
& \textbf{77.99}
& \textbf{79.80}
& 81.45
& \textbf{74.66} \\

w/o Short-term
& 58.02
& 77.31
& 78.20
& \textbf{81.77}
& 73.83 \\[0.35ex]

\hline

\multicolumn{6}{@{}l@{}}{\makebox[0pt][l]{%
\colorbox[gray]{0.95}{%
\makebox[\dimexpr\columnwidth-2\fboxsep\relax][c]{%
\rule[-0.3ex]{0pt}{2.1ex}\textit{Llama-3-8B}}}}} \\[0.3ex]

MixLoRA + PMR
& \textbf{76.96}
& \textbf{89.06}
& 85.40
& 85.64
& \textbf{84.26} \\

w/o Short-term
& 76.37
& 88.59
& \textbf{85.80}
& \textbf{86.07}
& 84.21 \\[0.35ex]

\hline

\multicolumn{6}{@{}l@{}}{\makebox[0pt][l]{%
\colorbox[gray]{0.95}{%
\makebox[\dimexpr\columnwidth-2\fboxsep\relax][c]{%
\rule[-0.3ex]{0pt}{2.1ex}\textit{Llama-3.1-8B-Base}}}}} \\[0.3ex]

MixLoRA + PMR
& \textbf{77.05}
& \textbf{88.43}
& \textbf{86.20}
& \textbf{86.83}
& \textbf{84.63} \\

w/o Short-term
& 76.88
& 87.37
& 82.40
& 86.40
& 83.26 \\[0.35ex]

\hline

\multicolumn{6}{@{}l@{}}{\makebox[0pt][l]{%
\colorbox[gray]{0.95}{%
\makebox[\dimexpr\columnwidth-2\fboxsep\relax][c]{%
\rule[-0.3ex]{0pt}{2.1ex}\textit{Qwen2.5-7B}}}}} \\[0.3ex]

MixLoRA + PMR
& \textbf{87.03}
& 92.05
& \textbf{92.80}
& \textbf{88.41}
& \textbf{90.07} \\

w/o Short-term
& 86.69
& \textbf{93.10}
& 92.00
& 88.08
& 89.97 \\[0.35ex]

\hline

\multicolumn{6}{@{}l@{}}{\makebox[0pt][l]{%
\colorbox[gray]{0.95}{%
\makebox[\dimexpr\columnwidth-2\fboxsep\relax][c]{%
\rule[-0.3ex]{0pt}{2.1ex}\textit{Qwen3-8B-Base}}}}} \\[0.3ex]

MixLoRA + PMR
& \textbf{89.93}
& \textbf{93.35}
& \textbf{93.20}
& \textbf{89.72}
& \textbf{91.55} \\

w/o Short-term
& 88.99
& 92.97
& 92.60
& 88.68
& 90.81 \\[0.35ex]

\noalign{\hrule height 0.8pt}
\end{tabular*}
\endgroup
\vspace{-12pt}
\end{table}

\begin{figure*}[!t]
    \centering
    \includegraphics[width=\textwidth]
    {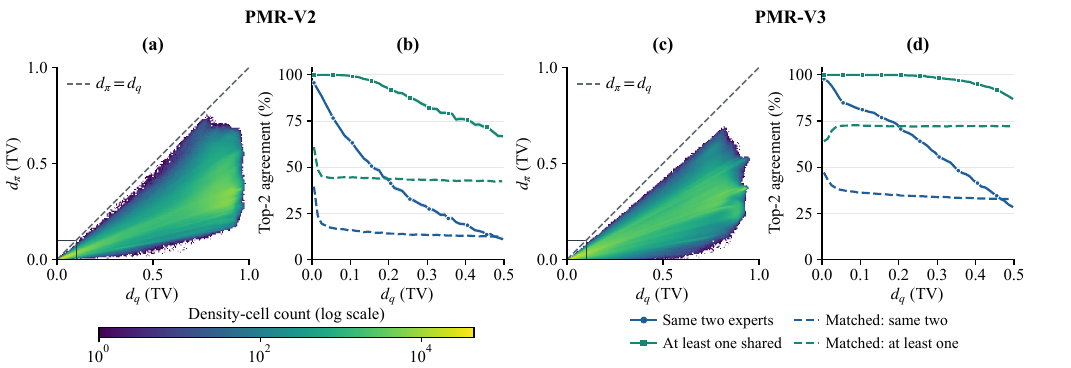}
    \caption{Prototype-response distances and Top-2 routing agreement in PMR-V2 (a,b) and PMR-V3 (c,d). (a,c): Density of token-pair TV-distance pairs \((d_q,d_\pi)\), where \(d_q\) and \(d_\pi\) measure differences in prototype-response distributions and expert probability distributions, respectively. The color scale indicates token-pair counts on a logarithmic scale, and dashed lines mark \(d_\pi=d_q\). (b,d): Proportions of token pairs selecting the same Top-2 experts (blue) or sharing at least one expert (green), plotted against \(d_q\). Dashed curves show matched baselines.}
    \label{fig:PMR-similarity}
\end{figure*}

\subsection{Analysis of Routing Components}
\label{sec:ablations}









To validate the contributions of PMR's key design components, we perform ablation studies and investigate the routing architecture, congestion pricing, and short-term sequence states.


\subsubsection{Routing Architecture}
We compare the linear--softmax router with the PMR router on TinyMixtral-V1, with congestion pricing disabled in PMR. Both configurations use the same auxiliary load-balancing loss~\cite{lepikhin2021gshard,fedus2022switch}, expert architecture, and Top-2 expert execution, under identical training and evaluation protocols. As reported in \cref{tab:PMR_routing_ablation}, PMR increases the seven-task mean zero-shot accuracy from 37.83 to 38.78 (+0.95 pp), where the large gains are observed on ARC-Easy (+3.80 pp), LAMBADA (+1.95 pp), and PIQA (+1.32 pp). These results support the effectiveness of PMR's routing architecture.

\subsubsection{Congestion Pricing Mechanism}
We then replace the auxiliary load-balancing loss with congestion pricing while retaining the PMR routing architecture. Tabulated in \cref{tab:PMR_routing_ablation}, replacing the auxiliary loss with congestion pricing raises the mean zero-shot accuracy from 38.78 to 39.37 (+0.59 pp), where the large gains occur on ARC-Easy (+2.91 pp) and OpenBookQA (+2.60 pp). These results support congestion pricing as an effective alternative to auxiliary load-balancing losses for regulating expert utilization and balancing token assignments, leading to improved task performance.

\subsubsection{Short-Term Sequence State}
\label{sec:PMR-state-ablation}



Short-term sequence state incorporates sequence-dependent information into prototype responses. To validate its contribution, we compare MixLoRA with its PMR-enhanced variant against a version that removes the Short-Term Sequence State while retaining the two-stage router, following the same fine-tuning and evaluation protocols as the main experiments. As reported in \cref{tab:PMR-state-ablation}, retaining the Short-Term Sequence State improves the mean accuracy across all five backbones, with gains of 0.83, 0.05, 1.37, 0.10, and 0.74 pp, respectively. Llama-3.1-8B-Base and Qwen3-8B-Base improve on all four benchmarks, with the largest gain reaching 3.80 pp on OpenBookQA for Llama-3.1-8B-Base. These findings establish short-term sequence state as an effective and robust means of improving the fine-tuning performance.


\begin{figure*}[t]
    \centering
    \includegraphics[width=\textwidth]
    {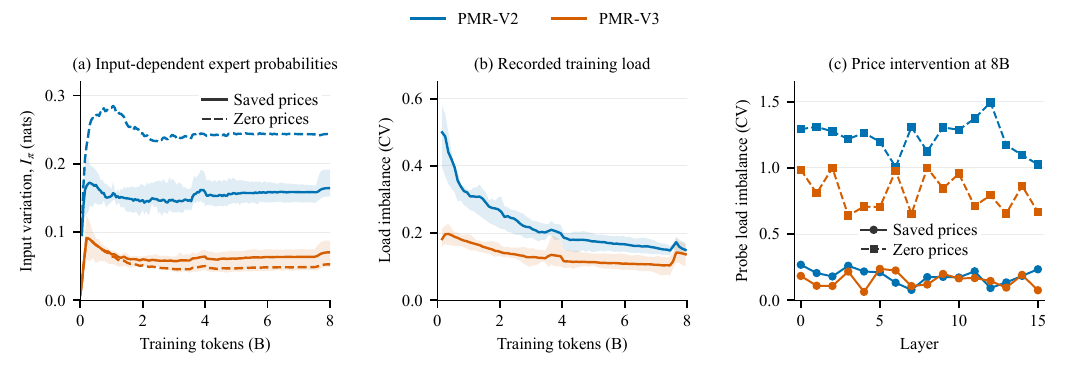}
    \caption{Input-dependent allocation diversity and load regulation in TinyMixtral-V2 and TinyMixtral-V3 with PMR routing. (a) Input-dependent distribution divergence \(I_\pi\) during training and at the final checkpoints, comparing saved congestion prices with zero prices. (b) Training dispatch coefficient of variation (CV), averaged across layers. (c) Final dispatch CV by layer on the two fixed probe sequences, comparing saved and zero prices. Shaded bands indicate interquartile ranges across layers for saved-price measurements.}
    \label{fig:allocation-diagnostics}
\end{figure*}

\subsection{Analysis of Routing Behavior}
\label{sec:PMR2-dynamics}
We further examine PMR's routing behavior from three perspectives: consistency across similar prototype distributions, diversity across prototypes and inputs, and load regulation through congestion pricing.

\subsubsection{Routing Consistency}
\label{sec:routing-similarity}

PMR maps each token's prototype distribution \(q_t\) to expert probabilities \(\pi_t=B^\top q_t\), from which the Top-2 experts are selected. We examine whether tokens with similar prototype distributions also receive similar expert probabilities and select similar experts. We use 128 text segments, with 32 each from BoolQ~\cite{clark2019boolq}, LAMBADA, HellaSwag, and ARC-Challenge. Both final checkpoints process the same token sequences, with association parameters and congestion prices fixed while short-term sequence states update causally. Across 16 layers, we compare all pairs of distinct token positions within each text, yielding 8,299,232 pairs per model. We use total variation (TV) distance~\cite{cpt_tv_levin2017} to measure differences between token pairs, denoting the distances between their prototype distributions and expert probability distributions by \(d_q\) and \(d_\pi\), respectively. \cref{fig:PMR-similarity}(a) and \cref{fig:PMR-similarity}(c) show that smaller \(d_q\) values are generally associated with smaller \(d_\pi\) values in both PMR-V2 and PMR-V3. This indicates that tokens with similar prototype distributions tend to receive similar expert probabilities. In addition, \(d_\pi\) remains below \(d_q\) across the observed range, with the gap becoming more pronounced at larger \(d_q\), particularly for PMR-V3.

We next examine whether tokens with similar prototype distributions also tend to select similar Top-2 expert sets. Within each prototype-distribution distance interval, we measure the proportion of token pairs selecting the same two experts, regardless of order, and the proportion sharing at least one expert.  Depicted in \cref{fig:PMR-similarity}(b) and \cref{fig:PMR-similarity}(d), we also provide the matched baselines (dashed curves), where reference agreement rates are computed from actual expert selections by matching token pairs within the same text and layer on token-ID equality and position-gap range, therefore accounting for differences in background overlap between models. For pairs with similar prototype distributions, exact Top-2 agreement substantially exceeds the matched baselines in both models, while nearly all pairs share at least one expert. As \(d_q\) increases, exact Top-2 agreement generally decreases and eventually falls below the baselines. These findings show that similar prototype distributions yield similar expert probabilities and overlapping Top-2 selections.

\subsubsection{Allocation Diversity}
\label{sec:PMR-conditional-allocation}


We further examine whether expert allocation varies across prototypes and input tokens. To quantify these differences, we use generalized Jensen--Shannon divergence~\cite{cpt2lin}. At the prototype level, \(D_{\mathrm{proto}}(B)\) measures divergence among prototype-level expert distributions represented by the rows of \(B\), weighting all prototypes equally. At the input level, \(I_\pi\) measures divergence among token-level expert distributions on two fixed 1,024-token probe sequences, weighting tokens equally within each layer. Both measures use natural logarithms and are zero when the distributions being compared are identical.

As shown by the solid curves in \cref{fig:allocation-diagnostics}(a), \(I_\pi\) initially increases, then declines and stabilizes during later training. Nevertheless, it remains positive, indicating that expert probabilities continue to vary across input tokens. At 8B tokens, mean \(I_\pi\) is 0.164 nats for PMR-V2 and 0.0702 nats for PMR-V3. Note that the corresponding mean \(D_{\mathrm{proto}}(B)\) values are 0.326 and 0.207 nats, respectively. Thus, expert allocation retains variation at both levels, with prototypes exhibiting distinct expert preferences and input tokens receiving different expert probability distributions.

\begin{figure*}[!t]
    \centering
    \includegraphics[width=\textwidth]
    {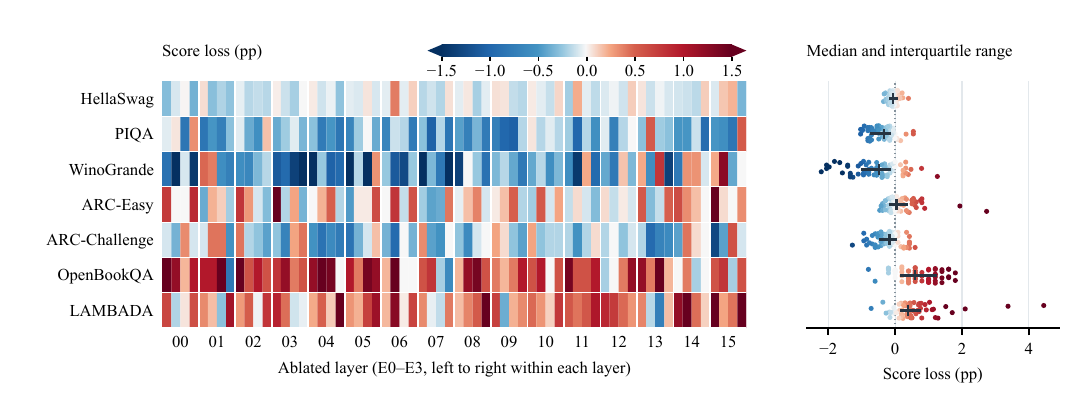}
    \caption{Heatmap of expert contributions in TinyMixtral-V3 with PMR routing after 8B training tokens, measured by the accuracy difference between the full model and the model with the corresponding expert removed. Results are shown across the seven pretraining benchmarks and 64 expert interventions, covering all four experts in each of the 16 layers. Layers are zero-indexed, with experts E0--E3 ordered from left to right within each layer. Positive values indicate accuracy degradation after expert removal.}
    \label{fig:expert-ablation}
\end{figure*}

\subsubsection{Load Balancing}
\label{sec:joint-B}

Next, we examine how expert loads evolve during training and whether congestion prices contribute to load balance. We measure load imbalance using the coefficient of variation (CV)~\cite{cpt_cv_nist} of expert dispatch shares within each layer, where lower values indicate more balanced expert usage. As shown in \cref{fig:allocation-diagnostics}(b), CV decreases substantially during early training in both models, with the decline slowing during later training. The resulting low CV indicates that expert usage becomes increasingly balanced and remains relatively balanced as training proceeds.

To examine the role of congestion prices in load regulation, we remove them at the final checkpoints while keeping the learned association scores \(C\) and prototype distributions \(\mathbf q\) fixed. On the same two probe sequences, this intervention increases load CV in every layer of both models (\cref{fig:allocation-diagnostics}(c)), validating the positive contributions of congestion prices to balancing expert usage across layers. 

Having established that congestion prices improve aggregate load balance in both models, we further examine whether this improvement comes from making expert probabilities more uniform across inputs, potentially reducing input-dependent allocation diversity, or whether balanced expert usage can be achieved while retaining distinct routing patterns across inputs. We therefore examine how removing congestion prices affects \(I_\pi\). Without congestion prices, $I_\pi$ increases in PMR-V2 but decreases in PMR-V3 (\cref{fig:allocation-diagnostics}(a)). Thus, although congestion prices generally improve load balance, its effect on input-dependent allocation differences varies with the learned routing mapping. These findings show that congestion prices improve aggregate load balance without necessarily imposing a uniform change in input-dependent allocation diversity.

\subsection{Analysis of Expert Contributions}
\label{sec:expert-ablation}


We analyze expert contributions on PMR-V3 as it has the highest token-to-parameter ratio during training. We perform 64 interventions, each zeroing one expert's output in one layer whenever selected while retaining the original Top-2 selections and gate weights. Downstream states and routing decisions are recomputed after each intervention. We measure each expert's contribution as the loss between the full-model score and the score obtained after the corresponding expert intervention, across the seven pretraining benchmarks. As shown in~\cref{fig:expert-ablation}, experts generally improve scores on LAMBADA and OpenBookQA, and tend to yield degrades on PIQA and WinoGrande. In the meantime, experts are associated with small effects on HellaSwag, while produces mixed effects on ARC-Easy and ARC-Challenge. Hence, expert contributions vary with the evaluation task.

Depicted in ~\cref{fig:expert-ablation}, OpenBookQA results provide a useful reference for examining these task differences. Among the 13 experts that contribute negatively or neutrally to OpenBookQA, 12 improve accuracy scores on LAMBADA, while all these 13 experts contribute negatives on PIQA, and 12 of them degrade the performance on WinoGrande. This overlap suggests that some expert locations support contextual word prediction, whereas their contributions do not necessarily extend to knowledge-based answer selection, physical reasoning, and other commonsense tasks. Also, none of the six experts with positive contributions to PIQA also contributes positively to WinoGrande, and only one does so for HellaSwag. These patterns show that experts benefiting physical commonsense reasoning do not necessarily support other forms of commonsense reasoning, despite their shared multiple-choice evaluation format. Overall, PMR routing exhibits differentiated expert contributions across evaluation tasks.

\subsection{Analysis of Blockwise State Processing}
\label{Training Efficiency}

We evaluate the efficiency of blockwise state processing and its effects on routing behavior and predictive performance on PMR-V3. We compare the default blockwise implementation against an eager tokenwise implementation, which updates the short-term sequence states and recomputes prototype distributions at every token. Both implementations use the same trained model parameters and congestion prices. Blockwise processing achieves 40,047 tokens/s, compared with 1,835 tokens/s for the eager tokenwise implementation, yielding a 21.82 times speedup in end-to-end training throughput. To assess the effect of blockwise processing on routing and prediction behaviors, we further compare the two implementations in terms of their routing decisions and next-token prediction performance using FP32 teacher-forced inference on 128 randomly sampled text windows, each containing 2,048 tokens. Blockwise processing achieves a mean TV distance of only 0.000650 from the tokenwise reference, with 99.73\% exact Top-2 set agreement, while the corresponding perplexities are 22.9606 and 22.9646 for blockwise and tokenwise execution, respectively, indicating similar predictive performance. These findings demonstrate that blockwise state processing substantially improves computational efficiency while closely preserving routing behavior and predictive performance.

\section{Conclusion}
PMR separates input matching from expert allocation by mapping token responses over shared prototypes to expert probabilities through a learned conditional association table. On the evaluated inputs, tokens with similar prototype responses tended to select overlapping sets of experts. In a separate fixed-input comparison, the saved congestion prices yielded lower dispatch imbalance than zero prices at both analyzed checkpoints, with the association scores and prototype responses held fixed. With two active experts per token, the full router improved mean accuracy in the evaluated pretraining and fine-tuning settings. The two-pass blockwise implementation also increased training throughput relative to eager tokenwise state processing while yielding similar expert selections and perplexity.

\end{document}